# Color Image Compression Based On Wavelet Packet Best Tree

Prof. Dr. G. K. Kharate

*Principal, Matoshri College of Engineering and Research Centre,*
*Nashik – 422003, Maharashtra, India*

Dr. Mrs. V. H. Patil

*Professor, Department of Computer Engineering, University of Pune*

## Abstract

In Image Compression, the researchers' aim is to reduce the number of bits required to represent an image by removing the spatial and spectral redundancies. Recently discrete wavelet transform and wavelet packet has emerged as popular techniques for image compression. The wavelet transform is one of the major processing components of image compression. The result of the compression changes as per the basis and tap of the wavelet used. It is proposed that proper selection of mother wavelet on the basis of nature of images, improve the quality as well as compression ratio remarkably. We suggest the novel technique, which is based on wavelet packet best tree based on Threshold Entropy with enhanced run-length encoding. This method reduces the time complexity of wavelet packets decomposition as complete tree is not decomposed. Our algorithm selects the sub-bands, which include significant information based on threshold entropy. The enhanced run length encoding technique is suggested provides better results than RLE. The result when compared with JPEG-2000 proves to be better.

## 1. Introduction

In today's modern era, multimedia has tremendous impact on human lives. Image is one of the most important media contributing to multimedia. The unprocessed image heavily consumes very important resources of the system. And hence it is highly desirable that the image be processed, so that efficient storage, representation and transmission of it can be worked out. The processes involve one of the important processes- "Image Compression". Methods for digital image compression have been the subject of research over the past decade. Advances in Wavelet Transforms and Quantization methods have produced algorithms capable of surpassing image compression standards. The recent growth of data intensive multimedia based applications have not only sustained the need for more efficient ways to encode the signals and images but also have made compression of such signals central to storage and communication technology. In Image Compression, the researchers' aim is to reduce the number of bits needed to represent an image by removing the spatial and spectral redundancies. Image Compression method used may be Lossy or Lossless. As lossless image compression focuses on the quality of compressed image, the compression ratio achieved is very low. Hence, one cannot save the resources significantly by using lossless image compression. The image compression technique with compromising resultant image quality, without much notice of the viewer is the lossy image compression. The loss in the image quality is adding to the percentage compression, hence results in saving the resources

There are various methods of compressing still images and every method has three basic steps: Transformation, quantization and encoding.

The transformation transforms the data set into another equivalent data set. For image compression, it is desirable that the selection of transform should reduce the size of resultant data set as compared to source data set. Many mathematical transformations exist that transform a data set from one system of measurement into another. Some mathematical transformations have been invented for the sole purpose of data compression; selection of proper transform is one of the important factors in data compression scheme.

In the process of quantization, each sample is scaled by the quantization factor whereas in the process of thresholding all insignificant samples are eliminated. These two methods are responsible for introducing data loss and it degrades the quality.

The encoding phase of compression reduces the overall number of bits needed to represent the data set. An entropy encoder further compresses the quantized values to give better overall compression. This process removes the redundancy in the form of repetitive bits.

We suggest the novel technique, which is based on wavelet packet best tree based on threshold Entropy with enhanced run-length encoding. This method reduces the time complexity of wavelet packets decomposition as complete tree is not decomposed. Our algorithm selects the sub-bands, which include significant information based on Threshold entropy. The results when compared with JPEG-2000 prove to be better. The basic theme of the paper is the extraction of the information from the original image based on Human Visual System. By exploring Human Visual interaction characteristics carefully, the compression algorithm can discard information, which is irrelevant to human eye.

## 2. Today's Scenario

The International Standards Organization (ISO) has proposed the JPEG standard [2, 4, 5] for image compression. Each color component of still image is treated as a separate gray scale picture by JPEG. Although JPEG allows any color component separation, images are usually separated into Red, Green, and Blue (RGB) or Luminance (Y), with Blue and Red color differences (U = B − Y, V = R − Y). Separation into YUV color components allows the algorithm to take the advantages of human eyes' lower sensitivity to color information. For quantization, JPEG uses quantization matrices. JPEG allows a different quantization matrix to be specified for each color component [3]. Though the JPEG provides good results previously, it is not perfectly suited for modern multimedia applications because of blocking artifacts.

Wavelet theory and its application in image compression had been well developed over the past decade. The field of wavelets is still sufficiently new and further advancements will continue to be reported in many areas. Many authors have contributed to the field to make it what it is today, with the most well known pioneer probably being Ingrid Daubechies. Other researchers whose contribution directly influence this work include Stephane Mallat for the pyramid filtering algorithm, and the team of R. R. Coifman, Y. Meyer, and M. V. Wickerhauser for their introduction of wavelet packet [6].

Further research has been done on still image compression and JPEG-2000 standard is established in 1992 and work on JPEG-2000 for coding of still images has been completed at end of year 2000. The JPEG-2000 standard employs wavelet for compression due to its merits in terms of scalability, localization and energy concentration [6, 7]. It also provides the user with many options to choose to achieve further compression. JPEG-2000 standard supports decomposition of all the sub-bands at each level and hence requires full decomposition at a certain level. The compressed images look slightly washed-out, with less brilliant color. This problem appears to be worse in JPEG than in JPEG-2000 [9]. Both JPEG-2000 and JPEG operate in spectral domain, trying to represent the image as a sum of smooth oscillating waves. JPEG-2000 suffers from ringing and blurring artifacts. [9]

Most of the researchers have worked on this problem and have suggested the different techniques that minimize the said problem against the compromise for compression ratio.

## 3. Wavelet And Wavelet Packet

In order to represent complex signals efficiently, a basis function should be localized in both time and frequency domains. The wavelet function is localized in time domain as well as in frequency domain, and it is a function of variable parameters.

The wavelet decomposes the image, and generates four different horizontal frequencies and vertical frequencies outputs. These outputs are referred as approximation, horizontal detail, vertical detail, and diagonal detail. The approximation contains low frequency horizontal and vertical components of the image. The decomposition procedure is repeated on the approximation sub-band to generate the next level of the decomposition, and so on. It is leading to well known pyramidal decomposition tree. Wavelets with many vanishing yield sparse decomposition of piece wise smooth surface; therefore they provide a very appropriate tool to compactly code smooth images. Wavelets however, are ill suited to represent oscillatory patterns [13, 14]. A special from a texture, oscillating variations, rapid variations in the intensity can only be described by the small-scale wavelet coefficients. Unfortunately, these small-scale coefficients carry very little energy, and are often quantized to zero even at high bit rate.

The weakness of wavelet transform is overcome by new transform method, which is based on the wavelet transform and known as wavelet packets. Wavelet packets are better able to represent the high frequency information [11].

Wavelet packets represent a generalization of multi-resolution decomposition. In the wavelet packets decomposition, the recursive procedure is applied to the coarse scale approximation along with horizontal detail, vertical detail, and diagonal detail, which leads to a complete binary tree. The pyramid structure of wavelet decomposition up to third level is shown in figure 4.1, tree structure of wavelet decomposition up to third level is shown in figure 4.2, structure of three level decomposition of wavelet packet is shown in figure 4.3, and tree structure of wavelet packets decomposition up to third level is shown in figure 4.4.

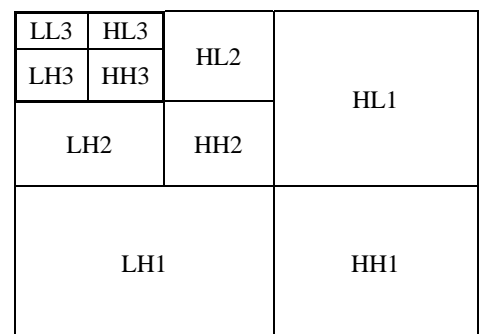

Figure 1 The-pyramid structure of wavelet decomposition up to third level

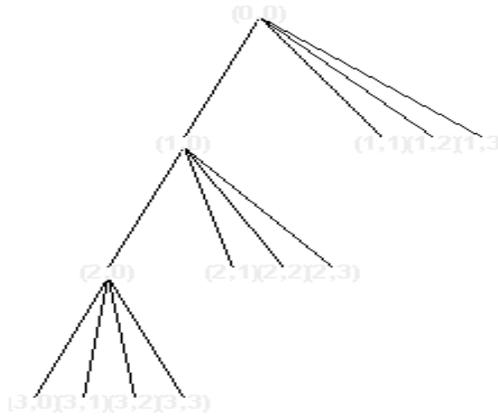

Figure 2 The-tree structure of wavelet decomposition up to third level

| LL₁LL₂ | LL₁HL₂ | HL₁LL₂ | HL₁HL₂ |
|---|---|---|---|
| $LL_1LH_2$ | $LL_1HH_2$ | $HL_1LH_2$ | $HL_1HH_2$ |
| $LH_1LL_2$ | $LH_1LH_2$ | $HH_1LL_2$ | $HH_1HL_2$ |
| $LH_1LH_2$ | $LH_1HH_2$ | $HH_1LH_2$ | $HH_1HH_2$ |

Figure 3 The structure of two level decomposition of wavelet packet

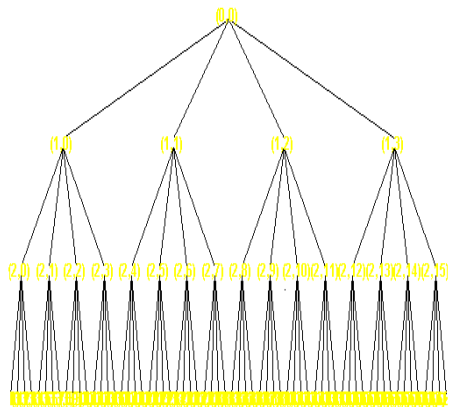

Figure 3 The complete decomposed three level tree

## 4. Proposed Algorithm for Image Compression

Modern image compression techniques use the wavelet transform for image compression. Considering the limitations of wavelet transform for Image Compression we suggest the novel technique, which is based on wavelet packet best tree based on Threshold Entropy with lossy enhanced run-length encoding. This method reduces the time complexity of wavelet packets decomposition, and selects the sub-bands, which include significant information in compact.

The Threshold entropy criterion finds the information contains of transform coefficients of sub-bands. Threshold entropy is obtained by the equation

$$Entropy = \sum_{i=0}^{N-1} abs(Xi) > Threshold \quad (1)$$

Where Xi is the i[th] coefficient of sub-band and N is the length of sub-band.

The information contains of decomposed components of wavelet packets may be greater than or less than the information contain of component, which has been decomposed. The sum of cost (Threshold entropy) of decomposed components (child nodes) is checked with cost of component, which has been decomposed (parent node). If sum of the cost of child nodes is less than the cost of parent node, then the child nodes are considered as leaf nodes of the tree, otherwise child nodes are neglected from the tree, and parent node becomes leaf node of the tree. This process is iterated up to the last level of decomposition.

The time complexity of proposed algorithm is less as compared to algorithm in paper [12]. In [12], the first wavelet packets decomposition of level 'J' takes place, and cost functions of all nodes in the decomposition tree are evaluated. Beginning at bottom of the tree, the cost function of the parent node is compared with union of cost functions of child nodes. According to the comparison of results the best basis node(s) is selected. This procedure is applied recursively at each level of the tree until the top most node of the tree is reached.

In proposed algorithm there is no need of full wavelet packets decomposition of level 'J' and no need to evaluate cost function of all nodes initially. Algorithm of best basis selection based on Threshold entropy is:

- Load the image
- Set current node equal to input image
- Decompose the current node using wavelet packet tree
- Evaluate the cost of current node, and decomposed components
- Compare the cost of parent node (current node) with the sum of cost of child nodes (decomposed components). If the sum of cost of child nodes is greater than the parent node, consider the parent node as leaf node of the tree, and child nodes are pruned, else repeat the steps 3, 4, and 5 for each child node by considering a child node as a current node, until last level of the tree reached.

This algorithm reduces the time complexity, because there is no need to decompose the full wavelet packets tree and no need to evaluate the costs initially. The decision of further decomposition, and cost calculation is based on the run time strategy of the algorithm, and it decides at run time whether to retain or prune the decomposed components [15].

Once the best basis has been selected based on cost function, the image is represented by a set of wavelet packets coefficients. The high compression ratio is achieved by using the thresholding to the wavelet packets coefficients. The advantages of wavelet packets can be gained by proper selection of thresholds.

Encoder further compresses the coefficients of wavelet packets tree to give better overall compression. Simple Run-Length Encoding (RLE) has proven very effective encoding in many applications. Run-Length Encoding is a pattern recognition scheme that searches for the repetition (redundancy) of identical data values in the code-stream. The data set can be compressed by replacing the repetitive sequence with a single data value and length of that data. We propose the modified technique for the encoding named as *Enhanced Run-Length Encoding,* and then for the bit coding well known Huffman coding or Arithmetic coding methods are used.

The problems with existing Run Length Encoding, is that the compression ratio obtained from run-length encoding schemes vary depending on the type of data to be encoded, and the repetitions present within the data set. Some data sets can be highly compressed by run-length encoding whereas other data sets can actually grow larger due to the encoding [16]. This problem of an existing run-length encoding techniques are eliminated up to the certain extent by using *Enhanced Run-Length Encoding* technique. In the proposed *Enhanced RLE,* the neighboring coefficients are compared, with acceptable value, which is provided by the user according to the applications. If the difference is less than the acceptable value then the changes are undone.

## 5. Results

The proposed algorithm is implemented and tested over the range of natural and synthetic images. The natural test images used are AISHWARYA, CHEETAH, LENA, BARBARA, MANDRILL, BIRD, ROSE, DONKEY, and synthetic images used are BUTTERFLY, HORIZONTAL, and VERTICAL. The results for 10 images is given in table1 and few output images are given in figure 4 are given as follows:

| Table 1 Results of selected Images | | | |
|---|---|---|---|
| **Image** | **Percentage of compression** | **Compression ratio** | **Peak signal to noise ratio (dB)** |
| AISHWARYA | 97.6873 | 50 | 64.8412 |
| CHEETAH | 97.3983 | 39 | 59.1732 |
| LENA | 98.4885 | 67 | 66.7228 |
| BARBARA | 97.0166 | 34 | 51.9716 |
| MANDRILL | 94.5585 | 19 | 46.1098 |
| BIRD | 97.1217 | 35 | 57.2633 |
| ROSE | 94.4339 | 18 | 46.7576 |
| DONKEY | 97.0665 | 35 | 50.6168 |
| BUTTERFLY | 96.0195 | 26 | 46.5698 |
| HORIZONTAL | 89.2611 | 10 | 47.7637 |
| VERTICAL | 97.5687 | 42 | 49.5102 |

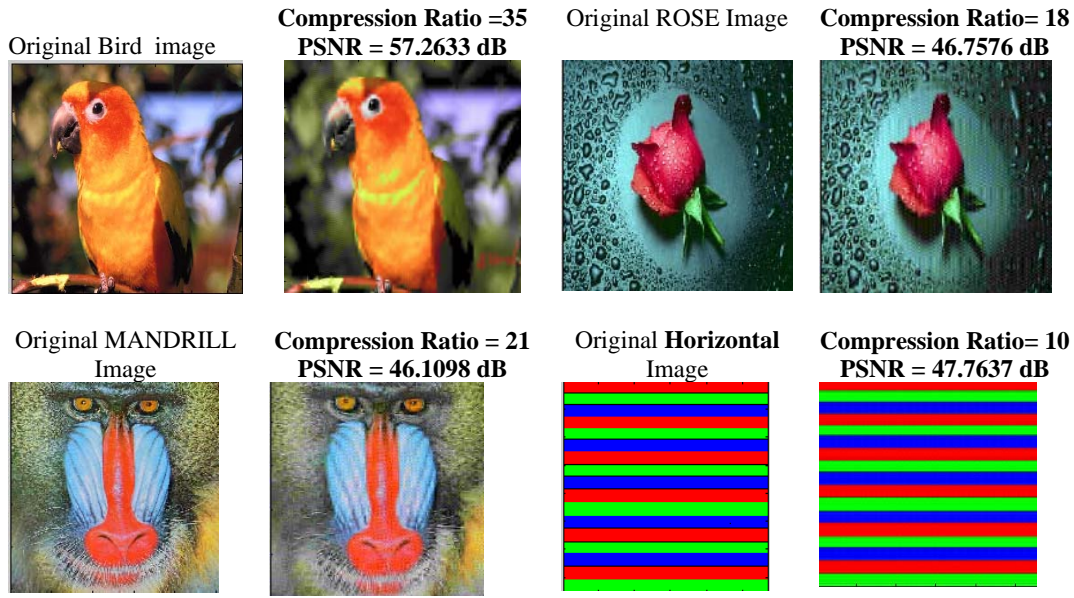

Fig. 4     Resultant Images with PSNR and compression ratio

## 6. Conclusion

The novel algorithm of image compression using wavelet packet best tree based on Threshold entropy and enhanced RLE is implemented, and tested over the set of natural and synthetic images and concluding remarks based on results are discussed. The results show that the compression ratio is good for low frequency (smooth) images, and it is observed that it is very high for gray images. For high frequency images such as Mandrill, Barbara, the compression ratio is good, and the quality of the images is retained too. These results are compared with JPEG-2000 application, and it is found that the results obtained by using the proposed algorithm are better than the JEPG2000.